%% file: iclr2026_conference.tex
\documentclass{article} 
\usepackage{iclr2026_conference,times}

\input{math_commands.tex}

\usepackage{hyperref}
\usepackage{url}
\usepackage{booktabs}
\usepackage{cleveref}
\usepackage{graphicx}
\usepackage{algorithm}
\usepackage{algorithmic}
\usepackage{tcolorbox}
\newtcolorbox{examplebox}[1][]{
  colback=blue!5!white,
  colframe=blue!60!black,
  fonttitle=\bfseries,
  title=Example,
  #1
}
\usepackage{amsmath}
\usepackage{amssymb}

\newcommand{\prompt}{\mathcal{P}}
\newcommand{\rules}{\mathcal{R}}

\newtheorem{definition}{Definition}
\newtheorem{observation}{Observation}

\title{ContraPrompt: Contrastive Prompt Optimization via Dyadic Reasoning Trace Analysis}

\author{%
  Rishav Rishav \quad Pushpak Pujari \quad Pushpendre Rastogi \\[0.3em]
  Vizops AI \\[0.2em]
  \texttt{\{rishav,pushpak,pushpendre\}@vizops.ai}%
}

\iclrfinalcopy
\begin{document}

\maketitle

\begin{abstract}
Prompt optimization methods either analyze individual failures in isolation or
compare prompt variants across examples, operating on single execution traces
with no access to the reasoning process that distinguishes success from failure
on the same input. We introduce \textbf{ContraPrompt}, built on the observation
that when a language model fails on a task but succeeds on a subsequent retry
with feedback, the difference between its two \emph{chain-of-thought traces}
constitutes an optimization signal not captured by prior prompt-optimization
methods operating on single traces or on final-output comparisons. Unlike
prior contrastive methods that compare final outputs or prompt variants, we
compare complete intermediate reasoning processes: the two traces share model,
input, and base prompt, so the differences that remain are the reasoning
strategy and (as a consequence of the retry mechanism) the appended error
feedback. We call this operation \emph{dyadic reasoning trace analysis}. The
multi-attempt solving phase is structured as an instrumented agentic retry
loop that generates this contrastive data automatically without human
annotation. Extracted rules are organized into an input-aware decision tree
that routes instructions by observable input characteristics. Evaluated on
four reasoning and compliance benchmarks, ContraPrompt outperforms
GEPA~\citep{agrawal2025gepa} on all four, with absolute gains of
$+8.29$ pp on HotPotQA ($+20.8\%$ rel.), $+2.21$ pp on GDPR-Bench
($+18.2\%$ rel.), $+7.14$ pp on GPQA~Diamond ($+10.6\%$ rel.), and
$+0.74$ pp on BBH ($+0.85\%$ rel.). Ablations confirm that dyadic trace
contrastivity is the critical component, with a $-16\%$ relative average
performance drop upon its removal. The mechanism generalizes beyond prompt
optimization: on 53 EvalSet black-box optimization problems, ContraPrompt beats
GEPA head-to-head on 11 problems, ties on 41, and loses on 1 at equal budget;
and on FiNER-139 financial named entity recognition~\citep{loukas2022finer}
(a 139-class high-cardinality classification task), ContraPrompt achieves
$+7.77$ pp over the unoptimized baseline ($+11.6\%$ rel.) and $+1.94$ pp
over GEPA ($+2.66\%$ rel.), with the input-aware tree producing branch
conditions that align with standard US GAAP financial-instrument categories.
We release artefacts such as optimized prompts for reproduction
here\footnote{\href{https://github.com/rishvv/contraprompt_artefacts/}{https://github.com/rishvv/contraprompt\_artefacts/}}.
\end{abstract}


\section{Introduction}
\label{sec:intro}

Agentic retry loops are now a standard component of LLM-based reasoning systems.
Reflexion~\citep{shinn2023reflexion} demonstrated that agents conditioned on
verbal reflections of their own prior failures improve substantially across
sequential decision-making, coding, and question answering. Self-Refine~\citep{madaan2023selfrefine}
showed that iterative self-feedback at inference time produces better outputs
across diverse generation tasks. SCoRe~\citep{kumar2024score} trained models
via reinforcement learning to internalize self-correction behavior, achieving
15.6\% absolute improvement on MATH. The shared premise is that LLMs benefit
from multiple attempts. Prior work on agentic retry loops has largely discarded
the failed-attempt traces after a successful retry; we show these traces
contain usable optimization signal when paired with the successful retry
on the same input.

Every successful retry produces a pair of chain-of-thought traces on the same
input: one that failed, and one that succeeded. Across four diverse benchmarks,
we observe that 20--37\% of first-attempt failures recover on retry with minimal
feedback, meaning that retry loops produce thousands of paired execution traces
as a byproduct. These pairs are structured comparisons: model identity, task
difficulty, domain, and base prompt are held fixed across the pair. The
differences that remain are the reasoning strategy applied on each attempt
and, because of how the retry loop works, the error feedback appended to the
second attempt. The second factor is an uncontrolled confound that we address
directly in the rule-extraction prompt (\cref{sec:extraction}) and discuss as
a limitation (\cref{sec:limitations}).

Existing prompt optimization methods do not exploit this signal. DPO~\citep{rafailov2023dpo}
constructs preference pairs from \emph{final outputs} and fine-tunes model
weights; it does not access intermediate reasoning. Contrastive prompt methods
such as those surveyed in~\citet{li2024lcp} compare prompt
\emph{variants}, not reasoning trajectories on the same input.
Failure-analysis methods~\citep{pryzant2023protegi,yuksekgonul2024textgrad}
read a single trace per failure and ask what went wrong, producing a negative
signal with no positive target. Evolutionary methods~\citep{agrawal2025gepa,yang2023opro}
compare prompt variants across examples, providing population-level signal but
no within-example process comparison. In all cases, the transition from failure
to success on the same input is treated as a byproduct of the retry mechanism
rather than as data.

We introduce \textbf{ContraPrompt}, which treats this transition as its primary
data source. The core operation is \emph{dyadic reasoning trace analysis}: given
two complete chain-of-thought traces $\tau^-$ and $\tau^+$ from the same model
on the same input (one failed, one successful), extract the reasoning delta
between them. Because the comparison is at the level of complete inference
chains rather than final answers, the extracted delta identifies a specific
reasoning step present in success and absent in failure, rather than describing
only what outputs to prefer. Extracted deltas are aggregated across the
training distribution, validated on held-out examples, and organized into an
input-aware decision tree that routes rules conditionally by observable input
features.

Our primary baseline is GEPA~\citep{agrawal2025gepa}. GEPA reads execution
trajectories monadically (one trace, one diagnosis) and proposes prompt
mutations along a cross-prompt, cross-example axis. DPO and contrastive-prompt
methods operate on final outputs or prompt phrasings, not on intermediate
reasoning steps. ContraPrompt reads trajectory \emph{pairs} from the same
input and extracts the within-example process delta, operating on a
within-prompt, within-example axis. These axes are orthogonal; a hybrid
combining GEPA's population-level search with ContraPrompt's instance-level
trace deltas is a natural direction for future work. For fair comparison, all
methods in this paper operate under a single-module constraint on the same
underlying model; GEPA's full HotPotQA pipeline uses a four-module architecture
and reaches different numbers that are not directly comparable (\cref{sec:setup}).

This paper makes three contributions. First, we motivate dyadic reasoning trace
analysis and ground it empirically as a source of step-level optimization signal
that final-output comparison does not provide (\cref{sec:theory}). Second, we
present the ContraPrompt system implementing this primitive via an instrumented
agentic retry loop, aggregated failure analysis, and input-aware tree-structured
rule organization (\cref{sec:method}). Third, we evaluate across four reasoning
and compliance benchmarks, black-box function optimization, and financial named
entity recognition, showing that the contrastive mechanism generalizes across
qualitatively different domains (\cref{sec:experiments,sec:extensions}). We
also observe that gains across the four reasoning benchmarks are ordered
consistently with each benchmark's retry success rate, which we report as a
suggestive pattern rather than a statistical finding given $n{=}4$
(\cref{sec:main-results}). Ablation studies and analysis are provided in
\cref{app:ablations,app:analysis}.

\begin{figure}[tbp]
\centering
\includegraphics[width=.95\textwidth]{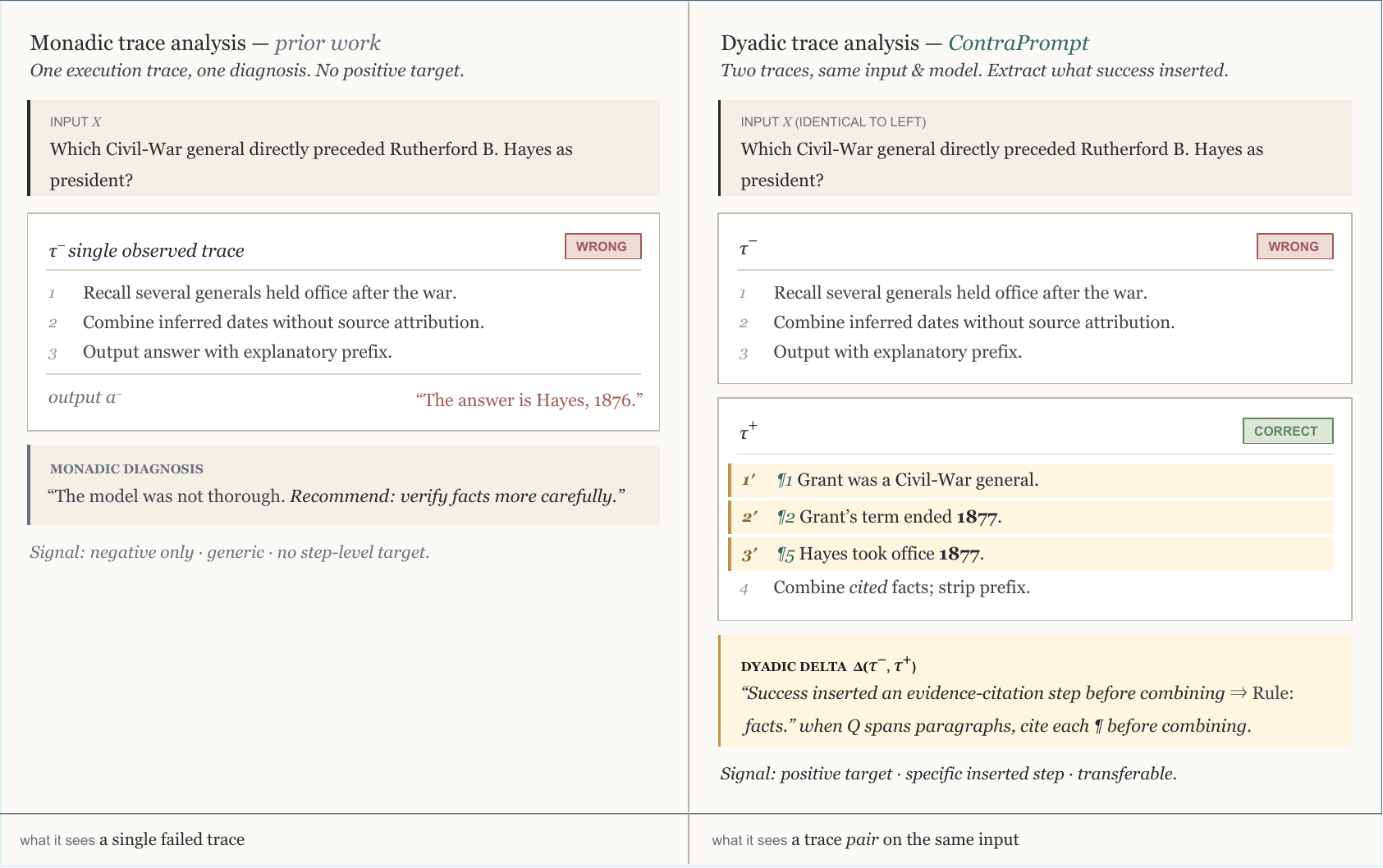}
\caption{Monadic vs. dyadic trace analysis. \emph{Left}: Prior methods consume one failed trace and produce a generic diagnosis with no positive target. \emph{Right:} ContraPrompt consumes a trace pair ($\tau^{-}$, $\tau^{+}$) produced by the same model on the same input, identifying the inserted reasoning steps that carry the success (highlighted, steps 1'--3'). The extracted rule targets that specific step, not generic thoroughness.}
\label{fig:monadic-vs-dyadic}
\end{figure}

\section{Related Work}
\label{sec:related}

\paragraph{Prompt optimization and monadic trace analysis.}
TextGrad~\citep{yuksekgonul2024textgrad} performs automatic differentiation through
text using LLM-generated feedback on individual outputs. ProTeGi~\citep{pryzant2023protegi}
diagnoses individual failures and proposes corrections via beam search over prompt
edits. Evolutionary approaches OPRO~\citep{yang2023opro} and APE~\citep{zhou2023ape}
search over populations of prompt candidates. All of these methods consume
execution traces monadically: one trace per diagnostic step. ContraPrompt
introduces a dyadic alternative in which the unit of analysis is a trace pair
from the same input.

\textbf{GEPA and the closest comparison.}
GEPA~\citep{agrawal2025gepa} is the current state of the art in reflective
prompt evolution and our primary baseline. GEPA reads execution trajectories
monadically to propose prompt mutations, maintaining a Pareto frontier of
candidates across example types. ContraPrompt reads trajectories dyadically,
extracting the process delta between failure and success on the same input.
These operate on orthogonal axes and are complementary.

\textbf{Agentic retry loops.}
Reflexion~\citep{shinn2023reflexion} generates verbal reflections on failed
episodes and conditions the next attempt on that stored reflection, consuming
one reflection within a single episode. ContraPrompt aggregates trace pairs
across the training set to extract rules that generalize to new inputs.
Self-Refine~\citep{madaan2023selfrefine} uses retry as an inference-time
quality mechanism. ContraPrompt inverts this priority: the comparison between
traces is the raw material for rule extraction, not the improved output itself.

\textbf{Contrastive and preference methods.}
DPO~\citep{rafailov2023dpo} constructs preference pairs from human annotations
on final answers and fine-tunes model weights. Contrastive prompting methods
such as those discussed in~\citet{li2024lcp} perform contrastive
analysis on prompt phrasings. Both compare final outputs or prompt inputs;
neither has access to intermediate reasoning processes. ContraPrompt compares
complete chain-of-thought traces, so extracted rules target specific reasoning
steps. SCoRe~\citep{kumar2024score} trains model weights via RL to improve
self-correction; ContraPrompt achieves related behavioral objectives by
extracting prompt rules from contrastive trace evidence, without any model
modification.

\textbf{Process-level signal and step-level supervision.}
Process reward models~\citep{lightman2023letsverify} and step-level
supervision methods train verifiers or reward models on intermediate reasoning
steps rather than final outputs. ContraPrompt is related in spirit---both
approaches extract signal from intermediate reasoning rather than from
answers---but differs in mechanism: we do not train a verifier, we extract
transferable textual rules from same-input failure/success pairs produced by
an unmodified model's retry loop.

\textbf{Memory and context engineering.}
Agentic Context Engineering (ACE)~\citep{zhang2026agenticcontextengineeringevolving}
organizes context strategically across long-horizon tasks via memory retrieval
and structured injection. Context engineering approaches motivate our input-aware
decision tree: just as memory systems gate retrieved context by relevance,
our tree routes extracted rules conditionally by input type, suppressing
irrelevant instructions and reducing noise. ContraPrompt's contribution is
the derivation of these conditional rules from contrastive trace pairs rather
than from retrieval signals.

\section{Motivation and Grounding}
\label{sec:theory}

\subsection{Notation}
\label{sec:notation}

Let $x$ denote a task input and $M$ a language model operating under base
prompt $\prompt$. A \emph{chain-of-thought trace} $\tau$ is the complete
sequence of intermediate reasoning steps produced by $M$ on input $x$ prior
to generating a final answer $a$. We write $\tau^+$ for a trace achieving a
higher task score and $\tau^-$ for one achieving a lower score on the same
input; correspondingly, $a^+$ and $a^-$ denote their final answers. The score
function $s(a, y)$ evaluates answer $a$ against ground truth $y$.

\subsection{The Capability-Application Gap}
\label{sec:cag}

\begin{definition}[Capability-Application Gap, empirical]
A model $M$ exhibits a capability-application gap on task $T$ if its
retry success rate
\[
\rho_T \;=\; P\bigl(s(\tau^+, y) > s(\tau^-, y) \,\big|\, \text{failed on first attempt}\bigr)
\]
under temperature 1.0 sampling is substantially larger than zero
(empirically, $\rho_T \gtrsim 0.2$ in our evaluation). This is an empirical
characterization of the task-model pair rather than a formal condition.
\end{definition}

A nonzero retry success rate implies that at least some first-attempt failures
are not capability deficits: the model can produce the correct reasoning, but
does not do so reliably on first attempt. This is the prerequisite for the
contrastive mechanism to function. Across our benchmarks, we observe $\rho_T$
values of approximately 37\% (HotPotQA), 25\% (GDPR-Bench), 24\% (GPQA~Diamond),
and 20\% (BBH). The temperature setting of 1.0 for task solving is deliberate:
at lower temperatures, retries converge toward the same deterministic output,
collapsing contrastive pairs to near-zero delta. At temperature 1.0, each retry
samples a different region of the model's reasoning strategy distribution, so
a successful second attempt more likely represents a genuinely distinct strategy.

\subsection{Why Reasoning Traces, Not Final Outputs}
\label{sec:info-advantage}

The central design choice of ContraPrompt is to compare complete chain-of-thought
traces rather than final outputs.

\begin{observation}[supported empirically in \cref{app:ablations}]
Dyadic comparison of reasoning traces yields step-level optimization signal
that dyadic comparison of final outputs alone does not provide.
\end{observation}

When only final outputs are compared (as in DPO or output-level contrastive
methods), one learns that answer $a^+$ was preferred over $a^-$ on input $x$.
What $M$ did differently to produce $a^+$ cannot be directly read off from
outputs alone. When complete chain-of-thought traces $\tau^+$ and $\tau^-$
are compared, the extractor can identify an inference step present in $\tau^+$
and absent in $\tau^-$---the evidence-attribution step, the unit-consistency
check, the explicit state enumeration---and target the extracted rule at that
step.

This is confirmed empirically in \cref{app:ablations}: replacing trace-level
comparison with answer-only comparison (using only final answers as input to
the rule extractor) reduces average performance by $-16\%$ relative, the same
degradation as removing contrastive mining entirely. The reasoning trace
carries information about the inference process that the final answer alone
does not.

\paragraph{What the pair does and does not control for.}
The pair $(\tau^-, \tau^+)$ holds model identity, task input, task difficulty,
domain, and base prompt fixed. It does \emph{not} hold conditioning context
fixed: the successful attempt is conditioned on error feedback from the prior
failure, which the failed attempt did not see. The pair is therefore a
two-trace observation of the same input under a within-subject design, not
a natural experiment isolating reasoning strategy. Rule extraction is prompted
to focus on reasoning-approach changes rather than conditioning changes
(\cref{sec:extraction}), but this decomposition is approximate; a controlled
comparison that retries without feedback at temperature 1.0 is a direction
for future work (\cref{sec:limitations}).

\subsection{Input-Aware Routing}
\label{sec:tree-theory}

As rules accumulate, injecting all of them into the prompt introduces
instruction noise: rules irrelevant to a given input type may confuse rather
than guide, and context length grows. The input-aware decision tree addresses
this by routing rules conditionally: for input $x$ with observable features
$\phi(x)$, the tree delivers only the subset $\mathcal{R}(x) \subseteq
\mathcal{R}_{\text{all}}$ relevant to $x$'s structure. This reduces expected
instruction length while preserving rule specificity per input type, paralleling
how memory and context engineering systems~\citep{zhang2026agenticcontextengineeringevolving}
gate retrieved content by relevance.

\section{Method}
\label{sec:method}

\begin{figure}[tbp]
\centering
\includegraphics[width=\textwidth]{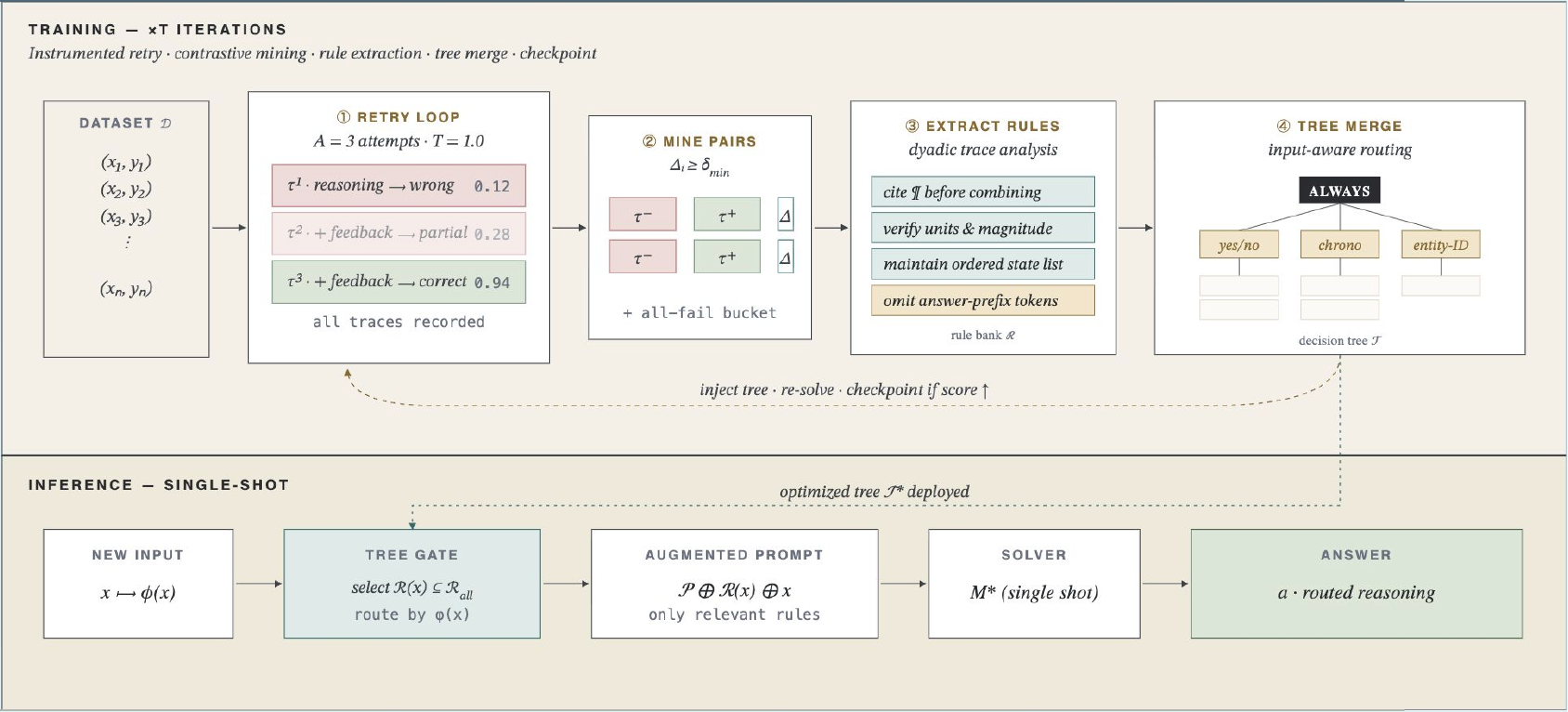}
\caption{\textbf{System Overview.}
Training (top): the instrumented retry loop produces contrastive trace pairs
$(\tau^{-}, \tau^{+})$ and an all-fail bucket; rules extracted dyadically feed
the input-aware tree, which is injected and checkpointed each outer iteration.
Inference (bottom): features $\varphi(x)$ route the input through the tree, so
the augmented prompt $\mathcal{P} \oplus \mathcal{R}(x) \oplus x$ carries only
the rule subset relevant to that input class.}
\label{fig:architecture}
\end{figure}

ContraPrompt optimizes prompts through an iterative loop with five phases:
instrumented agentic retry solving, contrastive pair mining, rule extraction
and aggregated failure analysis, input-aware tree merge, and checkpoint
evaluation. \Cref{fig:architecture} gives a high-level overview of the
system; the full procedure is given in \cref{alg:contraprompt} in the appendix.

\subsection{Instrumented Agentic Retry Loop}
\label{sec:multi-attempt}

For each training example, the model receives up to $A$ attempts (default $A{=}3$)
with error feedback from prior failures appended as additional context. On the
first attempt, the model sees only the original input under the base prompt
$\prompt$; on subsequent attempts, feedback from the prior failure is appended
as context. The base prompt $\prompt$ itself is never modified across attempts
within a single training example. All attempts elicit explicit chain-of-thought
reasoning traces; contrastive analysis always operates on complete reasoning
processes rather than final answers alone. Because attempts differ in both
reasoning strategy and appended feedback context, the pair $(\tau^-, \tau^+)$
is not a pure strategy comparison; the rule extractor is therefore prompted
to focus on the change in \emph{reasoning approach} rather than the change
in conditioning context (\cref{sec:extraction}).

Task solving uses temperature 1.0. At lower temperatures, retries tend to
reproduce the same failed reasoning with minor surface variation. At
temperature 1.0, each retry draws from a broader distribution of reasoning
strategies, increasing the probability that a successful retry represents a
genuinely distinct approach.

Feedback is calibrated to failure severity. For severe errors (score $< 0.3$),
feedback is coarse: ``Your previous answer was incorrect. Think more carefully.''
For partial failures (score $\geq 0.3$), feedback identifies the error type,
targeting the specific failure mode.

\subsection{Contrastive Mining}
\label{sec:mining}

We mine contrastive pairs by identifying examples where performance improved
substantially between attempts. For each example $x_i$, we compute the
per-example improvement $\Delta_i = c^+_i - c^-_i$, where $c^+_i =
\max_{j} s(a^j_i, y_i)$ is the score of the best attempt and $c^-_i =
\min_{j} s(a^j_i, y_i)$ the score of the worst attempt over all $A$ tries.
Pairs with $\Delta_i < \delta_{\min} = 0.02$ are discarded as noise. Remaining
pairs are ranked by $\Delta_i$ and used in subsequent rule extraction. Each
contrastive pair records the input, the failed reasoning trace $\tau^-$ with
its score, the successful reasoning trace $\tau^+$ with its score, and the
automatically inferred error type.

\subsection{Contrastive Rule Extraction}
\label{sec:extraction}

Given a contrastive pair $(\tau^-, \tau^+)$, a language model performs dyadic
trace analysis: it examines both complete traces, identifies specific reasoning
steps that differ, and extracts a transferable rule. The extraction prompt
presents both traces together with their appended feedback contexts and
instructs the extractor to focus on changes in reasoning approach rather than
changes in conditioning, asking: ``What did the improved reasoning do differently
in its \emph{chain of thought}? Extract a general rule that captures the
reasoning pattern.'' Rules follow the template:
``When [input pattern], [strategy] because [causal justification].''

The rule extractor operates on complete reasoning chains, not on final answers
alone. This enables extracted rules to target specific inference steps: the
extractor identifies a step that was present in success and absent in failure.
A subset of extracted rules target output formatting rather than reasoning
strategy (e.g., omitting answer prefixes for token-F1 benchmarks); such rules
are legitimate corrections to benchmark-relevant failure modes and are retained,
though they should not be conflated with reasoning-process changes. A taxonomy
distinguishing formatting corrections from reasoning changes is discussed in
\cref{sec:limitations}.

\subsection{Aggregated Failure Analysis}
\label{sec:failure-analysis}

Contrastive mining requires at least one successful retry and cannot produce
signal from examples where all attempts fail. For these cases, ContraPrompt
groups examples where all attempts failed by error type and analyzes each
group collectively to identify systematic patterns. Contrastive rules and
aggregated failure rules are complementary: the former describe what to change
when the model can succeed, and the latter describe what is structurally absent
when it cannot.

\subsection{Input-Aware Tree Merge}
\label{sec:tree-merge}

As rules accumulate, flat injection degrades performance through instruction
noise. The input-aware decision tree organizes rules into an \texttt{<always>}
section for universally applicable rules, and \texttt{<branch>} sections each
guarded by a condition observable from the input text. The tree is constructed
by a language model that clusters failing inputs by structural groupings and
assigns rules to relevant clusters. During tree construction, rules extracted
in the three-part template format are reformatted: the ``when'' clause becomes
the branch condition, and the ``because'' clause may be condensed to reduce
prompt length, preserving the actionable strategy component. Branch conditions
are restricted to features directly observable from the input, enabling
deterministic routing at inference time. The tree is constrained to two levels
of nesting.

On HotPotQA, branches correspond to yes/no questions, chronological
comparisons, entity identification, and multi-entity naming. On GPQA~Diamond,
branches correspond to organic reaction mechanisms, systematic enumeration,
structural assignment, quantitative calculations, and theoretical physics.
These decompositions emerge from the contrastive data without manual
specification.

\begin{examplebox}[title={Excerpt from Generated Rule Tree (HotPotQA)}]
\small
\texttt{<always>} \\
\quad \texttt{<rule>Output only the exact answer span without prefixes such as
``The answer is'', because token F1 scoring penalizes every extra token.</rule>} \\
\quad \texttt{<rule>Match the specificity level the question implies (species vs
genus, specific chemical vs category).</rule>} \\
\texttt{</always>} \\[4pt]
\texttt{<branch condition="Question asks yes/no structure">} \\
\quad \texttt{<rule>Return only ``yes'' or ``no'' with no additional text.</rule>} \\
\texttt{</branch>} \\[4pt]
\texttt{<branch condition="Question asks chronological comparison">} \\
\quad \texttt{<rule>Return only the entity name without dates or temporal
qualifiers.</rule>} \\
\texttt{</branch>}
\end{examplebox}

\section{Experimental Setup}
\label{sec:setup}

We evaluate ContraPrompt on benchmarks selected to span qualitatively distinct
failure regimes: multi-step evidence synthesis, structured reasoning over closed
categories, graduate-level scientific reasoning, and compliance classification.
This selection tests whether the dyadic mechanism generalizes across tasks where
first-attempt failures have different causal profiles: insufficient evidence
integration (HotPotQA), incorrect reasoning strategy (GPQA~Diamond), wrong
classification frame (GDPR-Bench), and structured task decomposition (BBH).

\textbf{Benchmarks.}
\textbf{HotPotQA}~\citep{yang2018hotpotqa} tests multi-hop question answering
requiring evidence synthesis across multiple passages (token F1, threshold 0.6).
\textbf{BBH}~\citep{suzgun2022bigbench} covers 23 BIG-Bench Hard tasks demanding
structured reasoning under exact match evaluation.
\textbf{GPQA~Diamond}~\citep{rein2023gpqa} tests graduate-level scientific
reasoning (accuracy, shuffled 4-option multiple choice), where errors typically
reflect misapplication of known principles.
\textbf{GDPR-Bench-Android}~\citep{ran2025gdprbench} requires multi-label
classification of GDPR compliance violations in Android app code (macro F1),
where failures frequently reflect applying the wrong regulatory frame.

\textbf{Models.}
Claude~Haiku~4.5 (\texttt{claude-haiku-4-5-20251001}) serves as the
task-solving model at temperature 1.0. Claude~Sonnet~4.5 serves as the rule
extraction and tree merge model at temperature 0.7. The two-model design
deliberately separates the distribution over which we measure first-attempt
performance from the model responsible for meta-level analysis, mirroring
standard practice in prompt optimization systems where a capable model
guides optimization of a leaner task solver.

\textbf{Task formulation and GEPA comparison.}
All benchmarks are formulated as single-module prompt optimization problems: one
language model call with one system prompt per input. This is a deliberate scope
choice: we are evaluating prompt optimization mechanisms, not system architectures.
GEPA's full HotPotQA pipeline uses a four-module architecture and reaches a
different operating point that is not directly comparable; that system answers
a different question. All methods (baseline, GEPA, ContraPrompt) operate under
the same single-module constraint on the same model, isolating the optimization
mechanism. Results should be interpreted accordingly.

Full experimental configuration details (data splits and ContraPrompt
hyperparameters) are provided in \cref{app:setup}.

\section{Results}
\label{sec:experiments}

\subsection{Main Results}
\label{sec:main-results}

\Cref{tab:main-results} presents the primary comparison. ContraPrompt outperforms
both baselines on all four benchmarks. Because the four benchmarks operate at
very different baseline levels (e.g., GEPA at 12.15 macro~F1 on GDPR-Bench vs.\
87.59 exact-match on BBH), we report both absolute (percentage-point) and
relative improvements.

\begin{table}[t]
\centering
\caption{Main results across four benchmarks. Best result per benchmark in
\textbf{bold}. Absolute improvements are in percentage points of the
corresponding metric; relative improvements are $(\text{ContraPrompt} -
\text{GEPA}) / \text{GEPA}$.}
\label{tab:main-results}
\vspace{0.5em}
\small
\begin{tabular}{@{}l l r r r r r@{}}
\toprule
Benchmark & Metric & Naive CoT & GEPA & \textbf{ContraPrompt} & $\Delta$ (abs.\ pp) & $\Delta$ (rel.) \\
\midrule
HotPotQA & Token F1 & 25.02 & 39.77 & \textbf{48.06} & $+8.29$ & $+20.8\%$ \\
BBH & Exact Match & 26.11 & 87.59 & \textbf{88.33} & $+0.74$ & $+0.85\%$ \\
GPQA Diamond & Accuracy & 63.27 & 67.35 & \textbf{74.49} & $+7.14$ & $+10.6\%$ \\
GDPR-Bench & Macro F1 & 10.12 & 12.15 & \textbf{14.36} & $+2.21$ & $+18.2\%$ \\
\bottomrule
\end{tabular}
\end{table}


The pattern of results is consistent with the capability-application gap
framing. HotPotQA has the highest retry success rate ($\approx$37\%); first-attempt
failures there predominantly reflect failure to connect evidence across
paragraphs rather than missing knowledge. Feedback redirects the model toward
explicit evidence citation, and the dyadic trace delta encodes this precisely:
failed traces skip source attribution; successful traces include it.
GPQA~Diamond's +10.6\% relative gain is consistent with reasoning failures
involving incorrect application of principles the model knows. GDPR-Bench's
+18.2\% relative gain suggests compliance failures frequently involve applying
the wrong classification frame, a failure mode amenable to contrastive
correction. BBH, where GEPA already achieves 87.59\%, shows a smaller $+0.74$
pp gain (+0.85\% relative), consistent with its lower retry success rate and
smaller pool of informative contrastive pairs.

\paragraph{Retry success rate and optimization effectiveness.}
Across these four benchmarks, ContraPrompt's improvement over GEPA is ordered
consistently with the retry success rate: HotPotQA (highest retry success,
largest gain) through BBH (lowest retry success, smallest gain). With $n{=}4$
benchmarks this is a suggestive pattern rather than a statistical finding,
and we do not report correlation statistics (any monotone ordering of four
points would yield a trivial rank correlation of 1.0). The ordering is
consistent with the prediction in \cref{sec:cag} that higher retry success
rates produce more and higher-quality contrastive pairs; testing this
hypothesis at $n$ large enough to support a statistical claim is a direction
for future work.

\section{Generalization to High-Cardinality Classification: FiNER-139}
\label{sec:finer}

The benchmarks in \cref{sec:experiments} involve a small set of output
categories and test primarily reasoning strategy. We now ask whether the
contrastive mechanism generalizes to a qualitatively different regime: 139-class
fine-grained domain classification where the challenge is precise disambiguation
among closely related categories, not reasoning strategy per se.

FiNER-139~\citep{loukas2022finer} requires classifying numerical values in SEC
filings into 139 XBRL taxonomy tags. The distinctions are subtle:
\texttt{LineOfCredit} covers the outstanding drawn amount, while
\texttt{LineOfCreditFacility\-Maximum\-BorrowingCapacity} covers only the
maximum available limit. These are invisible to any method operating at the
level of general reasoning instructions; the model must learn which feature
of the input value determines the correct tag within each financial instrument
class. We use DeepSeek-V3.1 as both task-solving and rule extraction model
to match the setup used in prior work on this benchmark by
\citet{zhang2026agenticcontextengineeringevolving}, enabling direct
comparability with that line of work.

Using 1{,}000 training examples, 500 validation examples, and 441 test examples,
ContraPrompt achieves 74.94\% accuracy: $+7.77$ pp absolute ($+11.6\%$ relative)
over the unoptimized baseline (67.17\%) and $+1.94$ pp absolute ($+2.66\%$
relative) over GEPA (73.0\%). Optimization converges in a single iteration
(730 seconds), producing 79 rules from contrastive pairs where the model
confused closely related tags.

The generated rule tree organizes these into 14 universal rules and 65
conditional rules across 11 domain-specific branches: debt instruments,
compensation and share-based payments, business combinations, lease agreements,
tax reconciliations, related party transactions, loss contingencies, intangible
assets, derivative instruments, segment reporting, and equity transactions.
These branch categories align with standard US GAAP financial-instrument
categories, reflecting the structure of the confusions present in the training
data. We note that this is a consequence of the training confusions clustering
by instrument type, not a claim that the method discovers GAAP independently;
still, the fact that the input-aware tree produces a categorization that
matches standard domain hierarchies is a useful organizational property.

\begin{examplebox}[title={Representative FiNER-139 Rules (Debt Instruments Branch)}]
\small
\texttt{For credit facilities, use LineOfCredit for outstanding amounts
and LineOfCreditFacility\-Maximum\-BorrowingCapacity only for the maximum
available limit.} \\[4pt]
\texttt{For interest expense specifically related to debt instruments,
prefer the more specific InterestExpenseDebt tag.} \\[4pt]
\texttt{For debt carrying amounts, use DebtInstrumentCarryingAmount,
not tags for face amount or maximum borrowing capacity.}
\end{examplebox}

The result demonstrates a property of the input-aware tree that the reasoning
benchmarks do not stress-test: its ability to scale to large numbers of
fine-grained conditional rules without interference. Flat injection of 79
rules into a prompt for a 139-class problem would create substantial noise;
the tree structure confines each disambiguation rule to the input context
where it is relevant, and universal rules handle cross-cutting failure modes
(e.g., \emph{``carefully distinguish between percentage values and monetary
values before selecting a tag''}).

\section{Generalization to Black-Box Function Optimization}
\label{sec:extensions}

We apply both ContraPrompt and GEPA \texttt{optimize\_anything}
\citep{agrawal2025gepa} to 53 synthetic test functions from the EvalSet
benchmark, spanning dimensions 1 to 11 and covering smooth polynomials, highly
multimodal landscapes, needle-in-haystack functions, and discontinuous
surfaces. Both methods operate as LLM-driven code evolution loops: starting
from a minimal random-search seed (a single uniform sample), each system
iteratively proposes improved Python solvers, executes them in a sandbox
against the objective function, and feeds structured diagnostics back to the
proposing LLM. In this setting, the \emph{dyadic primitive} operates on
optimizer evaluations rather than task reasoning traces: the worst- and
best-performing evaluations within each round form the contrastive pair,
and the rule extractor synthesizes a landscape summary from them.

\textbf{Setup.}
ContraPrompt uses a two-model architecture: Claude Haiku generates candidate
solutions and Claude Sonnet extracts contrastive strategy rules by comparing
worst- and best-performing evaluations within each run. Five evolution rounds
grow the solver from 9 lines to approximately 230 lines of structured code.
GEPA \texttt{optimize\_anything} uses a single reflection model (Claude Sonnet)
that proposes new solver code informed by Actionable Side Information (ASI),
including trial scores, tracebacks, and budget status. Both methods receive an
identical budget of 2{,}000 function evaluations per problem and are benchmarked
against Optuna (TPE sampler, same 2{,}000 evaluations).

\textbf{Results.}
In head-to-head comparison at equal budget, ContraPrompt wins on 11 problems,
ties on 41, and loses on 1 of the 53 shared problems. Against Optuna,
ContraPrompt records 21 wins, 30 ties, and 2 losses; GEPA records 14 wins,
30 ties, and 9 losses. ContraPrompt converges to the known global minimum
(gap $<10^{-4}$) on 41 of 53 problems, compared with 34 for GEPA. The high
tie counts against Optuna reflect that many EvalSet problems are easy enough
that multiple methods reach the same optimum; the more discriminating measure
is head-to-head losses, where ContraPrompt has 1 and GEPA has 9.

\textbf{Dimensional scaling.}
The gap widens with dimensionality. On low-dimensional problems ($d \leq 3$,
24 functions), both methods are near-identical: ContraPrompt leads 3 to 0 with
21 ties. On mid-range problems ($4 \leq d \leq 6$, 16 functions), ContraPrompt
leads 5 to 0 with 11 ties. On high-dimensional problems ($7 \leq d \leq 11$,
13 functions), ContraPrompt leads 3 to 1 with 9 ties. The most pronounced
failures for GEPA occur on Easom ($d{=}4,5$) and Styblinski-Tang ($d{=}5$),
where ContraPrompt matches global minima to full precision while GEPA incurs
gaps of 14.1 to 20.

\textbf{Mechanism of transfer.}
ContraPrompt's advantage traces to its explicit contrastive rule extraction
step. By pairing worst and best evaluations before each code synthesis round,
the reflection LLM receives a structured landscape summary (which regions
are dead zones, whether variables interact, whether the surface is smooth or
multimodal) rather than raw trial logs. This allows the LLM to rediscover
optimization techniques from contrastive rules alone, without being
provided algorithm names or templates. GEPA's single-step reflection over ASI
captures similar signals but without the explicit contrastive structure, which
appears to limit transfer on problems requiring multi-phase strategies. More
illustrative examples are in \cref{app:evalset}.

\section{Limitations and Future Work}
\label{sec:limitations}

\textbf{Dependence on nonzero retry success rate.}
The dyadic mechanism requires at least one successful retry per training example.
When all attempts fail, ContraPrompt falls back to aggregated failure analysis,
which provides weaker signal. The practical implication is that dyadic trace
analysis is most valuable where the capability-application gap is largest:
tasks where the model can succeed but does not do so reliably.

\textbf{Controlled comparison caveat.}
The contrastive pair $(\tau^-, \tau^+)$ shares model, input, and base prompt,
but the successful attempt is also conditioned on appended error feedback.
The pair is therefore not a pure reasoning-strategy comparison: the feedback
context is a second uncontrolled factor. Rule extraction is designed to focus
on reasoning-approach changes rather than conditioning changes, but this
decomposition is approximate. Disentangling these factors---for instance,
via a controlled experiment that retries without feedback at temperature 1.0
and compares the resulting reasoning delta to the fed-back condition---is
a direction for future work.

\textbf{Scope and scale.}
We evaluate on four primary benchmarks with a single task-solving model family.
Expanding to additional models, benchmarks, and training scales, and to
compound AI systems where failure may propagate across module boundaries,
are direct next steps.

\textbf{Future directions.}
Several design extensions are natural. A hybrid with GEPA's population-level
Pareto-frontier search would capture both within-example dyadic signal and
cross-example population signal, axes that are orthogonal and complementary.
An online variant updating the rule tree as new inputs arrive during deployment
would enable continual adaptation. Deeper or graph-structured rule organizations
may capture more complex input-type interactions.

\section{Conclusion}
\label{sec:conclusion}

\noindent Our central finding is that same-input failure-to-success trace pairs provide prompt-optimization signal that prior methods operating on single traces or on final outputs do not capture; the ablation further shows that replacing trace-level comparison with answer-only comparison produces the same $-16\%$ average relative degradation as removing contrastive mining entirely, indicating that the intermediate reasoning process, not the paired comparison structure per se, carries the signal. This translates into consistent absolute gains of $+8.29$ (HotPotQA), $+7.14$ (GPQA~Diamond), $+2.21$ (GDPR-Bench), and $+0.74$ pp (BBH) over GEPA, ordered consistently with each benchmark's retry success rate (a suggestive pattern at $n{=}4$); and the generalization to FiNER-139 and black-box function optimization shows that the dyadic primitive applies whenever a higher-quality and a lower-quality execution on the same problem can be compared to extract a transferable strategy rule. The input-aware decision tree, constructed automatically from these pairs, produces branch conditions that align with meaningful task decompositions---from multi-hop reasoning structure to US GAAP financial-instrument categories---because rules are grounded in the specific features that distinguish correct from incorrect reasoning on each input, and the tree organizes them accordingly.

\bibliography{iclr2026_conference}
\bibliographystyle{iclr2026_conference}

\appendix

\section{Full Algorithm}
\label{app:algorithm}

\begin{algorithm}[h]
\caption{ContraPrompt: Contrastive Prompt Optimization}
\label{alg:contraprompt}
\begin{algorithmic}[1]
\REQUIRE Training set $\mathcal{D}_{\text{train}}$, base module $M$, max
iterations $T$, max attempts $A$, patience $P$
\ENSURE Optimized module $M^*$
\STATE $\prompt \leftarrow \textsc{ExtractInstructions}(M)$; $\rules_{\text{all}}
\leftarrow \emptyset$; $s^* \leftarrow -1$; wait $\leftarrow 0$
\FOR{$t = 1$ to $T$}
    \STATE \COMMENT{\textbf{Phase 1: Instrumented agentic retry loop}}
    \STATE $\mathcal{A} \leftarrow \emptyset$
    \COMMENT{attempt records across all training examples}
    \FOR{each $x \in \mathcal{D}_{\text{train}}$}
        \STATE $\{(a^1_x, \tau^1_x), \ldots, (a^A_x, \tau^A_x)\} \leftarrow
        \textsc{SolveWithRetries}(M, x, A)$
        \STATE $\mathcal{A} \leftarrow \mathcal{A} \cup
        \{(x, \{(a^j_x, \tau^j_x)\}_{j=1}^{A})\}$
    \ENDFOR
    \STATE \COMMENT{\textbf{Phase 2: Contrastive pair mining (per-example)}}
    \STATE $\mathcal{C} \leftarrow \textsc{MineContrastivePairs}(\mathcal{A})$
    \COMMENT{$\Delta_i \geq \delta_{\min}$}
    \STATE \COMMENT{\textbf{Phase 3: Rule extraction and failure analysis}}
    \STATE $\rules_{\text{contrastive}} \leftarrow \emptyset$
    \FOR{each pair $(\tau^-, \tau^+) \in \mathcal{C}$}
        \STATE $r \leftarrow \textsc{ExtractRule}(\tau^-, \tau^+)$
        \COMMENT{Dyadic trace analysis}
        \STATE $\rules_{\text{contrastive}} \leftarrow \rules_{\text{contrastive}}
        \cup \{r\}$
    \ENDFOR
    \STATE $\rules_{\text{failure}} \leftarrow
    \textsc{AggregatedFailureAnalysis}(\text{all-fail examples in } \mathcal{A})$
    \STATE $\rules_{\text{all}} \leftarrow \rules_{\text{all}} \cup
    \rules_{\text{contrastive}} \cup \rules_{\text{failure}}$
    \STATE \COMMENT{\textbf{Phase 4: Input-aware tree merge}}
    \STATE $\mathcal{T} \leftarrow \textsc{TreeMerge}(\rules_{\text{all}},
    \text{failing inputs in } \mathcal{A})$
    \STATE \COMMENT{\textbf{Phase 5: Inject and checkpoint}}
    \STATE $M_t \leftarrow \textsc{InjectTree}(M, \mathcal{T})$
    \STATE $s \leftarrow \textsc{Evaluate}(\mathcal{D}_{\text{train}}, M_t)$
    \IF{$s > s^*$}
        \STATE $s^* \leftarrow s$; $M^* \leftarrow M_t$; wait $\leftarrow 0$
    \ELSE
        \STATE wait $\leftarrow$ wait $+ 1$
    \ENDIF
    \IF{wait $\geq P$}
        \STATE \textbf{break}
    \ENDIF
\ENDFOR
\RETURN $M^*$
\end{algorithmic}
\end{algorithm}

\section{Experimental Setup Details}
\label{app:setup}

\paragraph{Data splits.}
For HotPotQA, BBH, and GPQA~Diamond, we use 50 training and 50 validation
examples. Test evaluation uses 200 examples for HotPotQA, 540 for BBH, and 98
for GPQA~Diamond. For GDPR-Bench, we use 100 training, 100 validation, and 687
test samples.

\paragraph{ContraPrompt configuration.}
$A{=}3$ maximum attempts, up to $T{=}15$ outer iterations with patience $P{=}3$,
minimum improvement threshold $\delta_{\min}{=}0.02$, maximum two levels of
tree nesting. No benchmark-specific tuning.

\paragraph{GEPA comparison protocol.}
GEPA's full HotPotQA pipeline uses a four-module architecture and achieves 62.33
F1 on Qwen3-8B; we disclose this for context but that system answers a different
question. All methods (baseline, GEPA, ContraPrompt) operate under the same
single-module constraint on the same model.

\section{Ablation Studies}
\label{app:ablations}

\begin{figure}[h]
\centering
\includegraphics[width=0.56\textwidth]{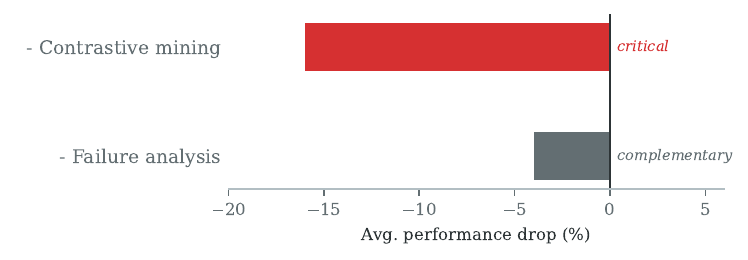}
\caption{Average relative performance drop when each component is removed.
Contrastive mining produces the largest degradation ($-16\%$ relative); the
input-aware tree and aggregated failure analysis provide smaller complementary
gains.}
\label{fig:ablation}
\end{figure}

\Cref{fig:ablation} presents component ablations, removing one component at a
time from the full system and measuring the average relative performance drop
across all four benchmarks. ``Average relative drop'' is computed as the mean,
across the four benchmarks, of $(\text{score}_{\text{full}} -
\text{score}_{\text{ablated}})/\text{score}_{\text{full}}$.

Removing multi-attempt solving and contrastive pair mining produces a $-16\%$
average relative drop. Without contrastive pairs, the method reduces to
failure-only analysis, which provides a negative signal (what to avoid) but
no positive target (what to do instead). The deficit is largest on HotPotQA
($-13.8$ percentage points of F1), precisely where retry success rate and
contrastive pair volume are highest. This is consistent with the central claim
that the dyadic trace comparison is the primary optimization signal.

Replacing the input-aware tree with flat rule injection produces a $-6\%$
average relative drop. The benefit is concentrated on benchmarks with
heterogeneous input types (HotPotQA, GPQA~Diamond), where irrelevant rules
interfere with processing of out-of-scope inputs.

Removing aggregated failure analysis produces a $-4\%$ average relative drop,
consistent with its role as a fallback for the capability-deficit regime
where contrastive pairs are unavailable.

\paragraph{Trace-level vs.\ answer-only extraction.}
To directly test whether reasoning traces carry signal beyond the pairing
structure, we ran a variant of the rule extractor that receives only final
answers rather than complete chain-of-thought traces. This variant matches the
performance of removing contrastive mining entirely ($-16\%$ average relative
drop), indicating that the reasoning trace, not merely the paired comparison
structure, is the source of the step-level optimization signal. This
distinguishes ContraPrompt from final-output contrastive methods such as
DPO and prompt-level contrastive approaches.

\section{Analysis: Interpretability of Extracted Rules}
\label{app:analysis}

The rules extracted by dyadic trace analysis target specific reasoning steps
by construction: each rule describes a reasoning step present in a successful
trace and absent in a failed trace on the same input.

\begin{figure}[h]
\centering
\includegraphics[width=\textwidth]{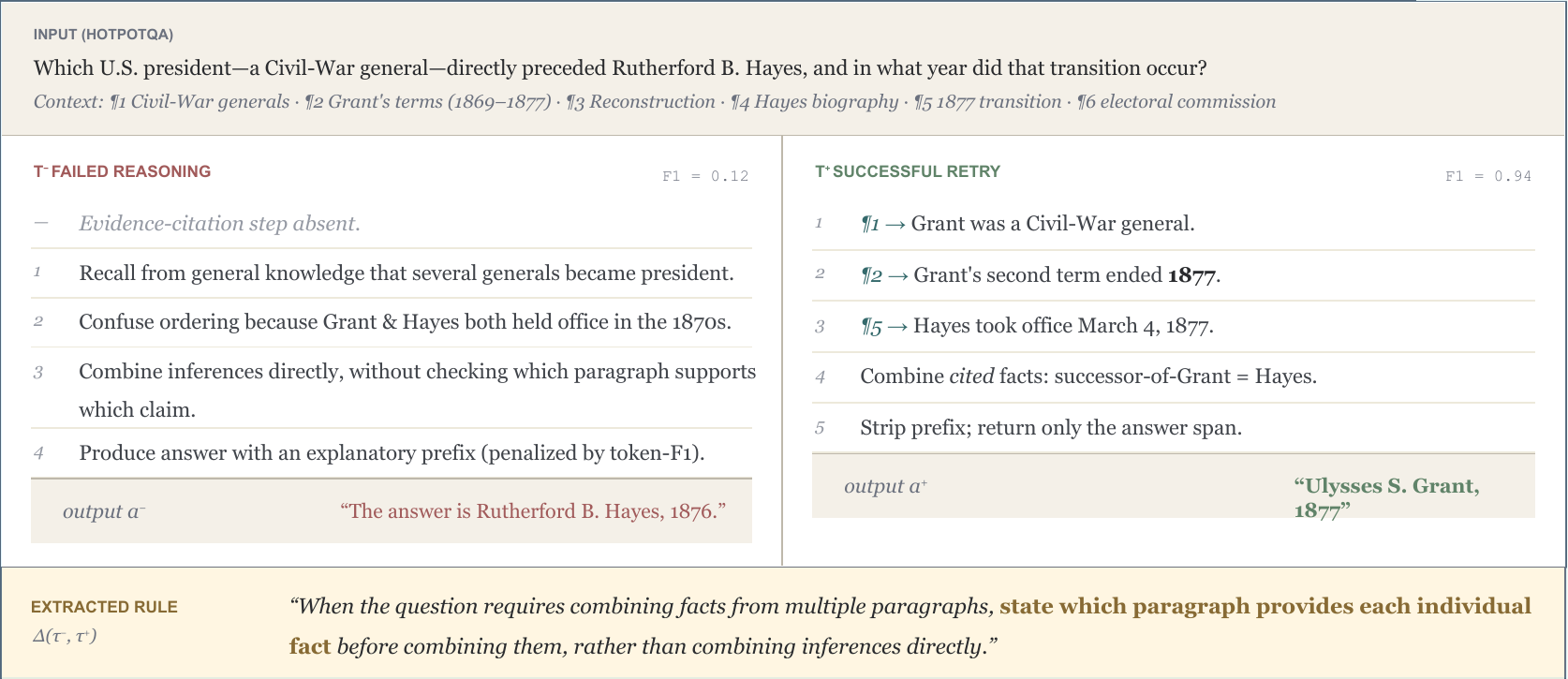}
\caption{A representative contrastive pair on HotPotQA. Both traces share model, input, and base prompt; they differ in reasoning strategy (and, as a consequence of the retry mechanism, the appended error feedback). The extracted rule targets the specific reasoning step that was inserted on retry—explicit paragraph-level attribution before combining facts—rather than the generic prescription "be more thorough".}
\label{fig:contrastive-example}
\end{figure}

Representative rules from three benchmarks illustrate the specificity achievable:

\begin{examplebox}[title={Representative Extracted Rules}]
\small
\textbf{HotPotQA:} ``When the question requires combining facts from multiple
paragraphs, state which paragraph provides each individual fact before combining
them, rather than combining inferences directly.'' \\[4pt]
\textbf{GPQA~Diamond:} ``When a question involves a quantitative physical
calculation, verify that the units and order of magnitude of the intermediate
result are consistent with the physical scenario before proceeding to the
final answer.'' \\[4pt]
\textbf{BBH (object tracking):} ``When the task involves tracking object
states through a sequence of swap operations, maintain an explicit ordered
state list that is updated after each individual operation rather than
reasoning about the final state directly.''
\end{examplebox}

Each rule names a specific step in the reasoning chain rather than offering
generic advice. This specificity follows from extracting rules from trace
comparisons rather than from monadic failure diagnoses. GEPA's monadic
reflection tends to produce higher-level prescriptions about thoroughness.
Both rule types are useful; the two methods are complementary.

\section{Black-Box Optimization: Illustrative Examples}
\label{app:evalset}

We highlight representative problems illustrating behavioral differences
between ContraPrompt, GEPA \texttt{optimize\_anything}, and Optuna. Complete
artifacts are available in our repository~\href{https://anonymous.4open.science/r/artefacts-F50C}{here}.

\paragraph{Easom (5D).}
The Easom function has a single narrow basin near $(\pi, \pi, \ldots, \pi)$
and is nearly flat across the rest of its domain. Optuna's 2{,}000 TPE trials
never locate this basin (gap 5.03), and GEPA fares worse (gap 15.5), exhausting
its budget on global exploration without concentrating search near the basin.
Contrastive analysis of best versus worst evaluations produces rules identifying
that improved solutions cluster near the center of the positive quadrant;
from these, the LLM synthesizes a solver that enumerates candidate regions
around $(\pi, \pi, \ldots, \pi)$, runs coordinate descent along each dimension
independently, and refines with L-BFGS-B from diverse starting points. The
final solver finds the exact global minimum (gap $< 10^{-15}$) in 194 lines
of Python. No human specified this strategy; the contrastive landscape summary
led the LLM to identify the basin location from evaluation pairs. This problem
exemplifies why structured landscape summaries outperform raw trial logs:
without contrastive pairs identifying the flat-versus-basin structure, the
reflection LLM has no signal to motivate targeted search.

\paragraph{McCourt10 (8D).}
Both ContraPrompt and GEPA converge to the exact global minimum of $-2.519$ to
12 decimal places (gap $< 5 \times 10^{-12}$), while Optuna stalls at a gap of
0.015. Both synthesized solvers follow a similar three-phase pipeline:
warm-start from prior best points, differential evolution with the best known
solution seeded into the initial population, and L-BFGS-B local refinement.
This is one of the problems where both code evolution approaches converge to
effectively identical strategies, suggesting that when the landscape provides
clear gradient signal, either feedback mechanism suffices.

\paragraph{McCourt11 (8D).}
The sole problem where GEPA outperforms ContraPrompt. GEPA reaches a gap of
$3.9 \times 10^{-8}$ from the global minimum ($f^* = -0.3905$), while
ContraPrompt stalls at a gap of 0.074. GEPA's winning solver chains Gaussian
process surrogate modeling with multi-start local optimization and dual
annealing, allocating budget adaptively based on improvement rate.
ContraPrompt's contrastive rules correctly identify the promising region but
the synthesized code underallocates budget to local refinement, suggesting
that on landscapes where surrogate-guided search is critical, GEPA's richer
ASI (which includes all prior trial coordinates) can outperform contrastive
summaries that compress this information into textual rules.

\paragraph{Ackley (11D).}
Contrastive rules surface that the function is approximately separable, leading
ContraPrompt to synthesize a six-phase pipeline leading with coordinate-wise
search before chaining Powell's method, Nelder-Mead, and L-BFGS-B. The final
gap of 3.26 improves on Optuna's 5.52 by 41\%. GEPA produces a gap of 15.1,
nearly $5\times$ worse, despite generating a longer solver. Without contrastive
rules identifying separability, the solver allocates most budget to
full-dimensional differential evolution, which is inefficient in 11 dimensions
at the 2{,}000-evaluation scale.

\paragraph{DeflectedCorrugatedSpring (4D).}
ContraPrompt finds the exact global minimum ($f^* = -1.0$, gap $< 10^{-13}$),
while both GEPA and Optuna converge to a local minimum at $f = -0.843$.
The chronology is as follows: early random search within the code evolution
loop encounters an evaluation near $-1.0$ by chance; this evaluation is
recorded alongside the dominant $-0.843$ local attractor evaluations. The
contrastive pair between $-0.843$ and $-1.0$ exposes the existence of the
deeper basin, prompting the LLM to synthesize a solver with explicit multi-basin
detection: it restarts from random initializations after each local convergence,
maintaining a diverse archive of local optima. The subsequent generalized solver
reliably relocates and refines the deep basin across restarts. Neither
GEPA's reflection nor Optuna's TPE sampler construct this multi-basin logic,
and both remain trapped at $-0.843$.

\end{document}

%% file: math_commands.tex
\usepackage{amsmath,amsfonts,bm}

\def\eqref#1{equation~\ref{#1}}

\def\1{\bm{1}}

\DeclareMathAlphabet{\mathsfit}{\encodingdefault}{\sfdefault}{m}{sl}
\SetMathAlphabet{\mathsfit}{bold}{\encodingdefault}{\sfdefault}{bx}{n}

